\documentclass[letterpaper]{article} % DO NOT CHANGE THIS
\usepackage[submission]{aaai24}  % DO NOT CHANGE THIS
\usepackage{times}  % DO NOT CHANGE THIS
\usepackage{helvet}  % DO NOT CHANGE THIS
\usepackage{courier}  % DO NOT CHANGE THIS
\usepackage[hyphens]{url}  % DO NOT CHANGE THIS
\usepackage{graphicx} % DO NOT CHANGE THIS
\usepackage{natbib}  % DO NOT CHANGE THIS AND DO NOT ADD ANY OPTIONS TO IT
\usepackage{caption} % DO NOT CHANGE THIS AND DO NOT ADD ANY OPTIONS TO IT
\usepackage{algorithm}
\usepackage{algorithmic}
\usepackage{booktabs}       % professional-quality tables
\usepackage{multirow}
\usepackage{makecell}
\usepackage{amsmath}

\usepackage{newfloat}
\usepackage{listings}
\DeclareCaptionStyle{ruled}{labelfont=normalfont,labelsep=colon,strut=off} % DO NOT CHANGE THIS
\floatstyle{ruled}
\newfloat{listing}{tb}{lst}{}
\floatname{listing}{Listing}
\makeatletter\def\showauthors@on{T}\makeatother
\title{ssProp: Energy-Efficient Training for Convolutional Neural Networks with \underline{S}cheduled \underline{S}parse Back \underline{Prop}agation}
\author{
    Lujia Zhong\textsuperscript{\rm 1,2},
    Shuo Huang\textsuperscript{\rm 1,3},
    Yonggang Shi\textsuperscript{\rm 1,2,3},
}
\affiliations{
    \textsuperscript{\rm 1}Stevens Neuroimaging and Informatics Institute, University of Southern California\\
    \textsuperscript{\rm 2} Ming Hsieh Department of Electrical and Computer Engineering, University of Southern California\\
    \textsuperscript{\rm 3} Alfred E. Mann Department of Biomedical Engineering, University of Southern California\\
}

\begin{document}

\maketitle

\begin{abstract}
Recently, deep learning has made remarkable strides, especially with generative modeling, such as large language models and probabilistic diffusion models. However, training these models often involves significant computational resources, requiring billions of petaFLOPs. This high resource consumption results in substantial energy usage and a large carbon footprint, raising critical environmental concerns.
Back-propagation (BP) is a major source of computational expense during training deep learning models. To advance research on energy-efficient training and allow for sparse learning on any machine and device, we propose a general, energy-efficient convolution module that can be seamlessly integrated into any deep learning architecture. Specifically, we introduce channel-wise sparsity with additional gradient selection schedulers during backward based on the assumption that BP is often dense and inefficient, which can lead to over-fitting and high computational consumption.
Our experiments demonstrate that our approach reduces 40\% computations while potentially improving model performance, validated on image classification and generation tasks. This reduction can lead to significant energy savings and a lower carbon footprint during the research and development phases of large-scale AI systems. Additionally, our method mitigates over-fitting in a manner distinct from Dropout, allowing it to be combined with Dropout to further enhance model performance and reduce computational resource usage. Extensive experiments validate that our method generalizes to a variety of datasets and tasks and is compatible with a wide range of deep learning architectures and modules. Code is publicly available at https://github.com/lujiazho/ssProp.
\end{abstract}

\section{Introduction}

In recent years, deep learning and neural networks have advanced rapidly, bringing significant benefits but also raising concerns about energy consumption and environmental impact. While deep learning research communities primarily focus on novel technologies and enhancing performance, the significant concerns of energy consumption and carbon footprint are often overlooked. 
According to the Stanford AI Index Report 2024 \cite{stanfordaiindex2024}, training leading AI models across language and vision-language fields \cite{achiam2023gpt,anil2023gemini,liu2024sora,alayrac2022flamingo,esser2024scaling} can cost millions of dollars and require over 10 billion petaFLOPs ($1$ petaFLOPs $\approx 10^{15}$ FLOPs, FLOPs: floating-point operations). This immense computational need results in high power and cooling requirements, contributing to a growing carbon footprint \cite{kuo2023green,de2023growing,lannelongue2021green,schwartz2020green,wu2022sustainable,xu2021survey}.

Sparsification is a major technique to address this challenge, distinct from methods that reduce training costs by speeding up model convergence, such as initialization \cite{he2015delving,glorot2010understanding} and normalization \cite{ioffe2015batch,wu2018group}. Previous research has explored various strategies of sparsification. MeProp \cite{sun2017meprop} significantly reduces computational costs by computing only a subset of gradients, yet its application is limited to multi-layer perceptrons (MLPs). MeProp-CNN \cite{wei2017minimal,sun2018training} extends this approach to convolutional neural networks (CNNs) with a different design for gradient sparsification from ours. 
Another approach, presented by Wang wt al. \cite{wang2019accelerated}, approximates backward gradient maps with a top-k selection method. 
MSBP \cite{zhang2020memorized} and Resprop \cite{goli2020resprop} apply gradient sparsification to accelerate training by reusing calculated gradients for future updates. 
SWAT \cite{raihan2020sparse} sparsifies both forward and backward passes in a simulated environment with minimal accuracy degradation, yet it is restricted to A100 or equivalent GPUs that support hardware sparsification acceleration. 
Additionally, \cite{ye2020accelerating} proposes a gradient pruning method, demonstrating training speed-up on CPUs with minor accuracy loss. 
\cite{zhou2021efficient} achieves sparse forward and backward by formulating sparsification as an optimization problem and estimating gradient with a guaranteed variance bound, requiring two forward passes for each update iteration. 
These works have seen some success but suffer from one or more of the following limitations: insufficient evaluation (validated on small datasets, e.g., MNIST, CIFAR-10), compromised accuracy, complex implementation, limited generality, and reliance on specific hardware. Besides, most approaches focus on speeding up training by sparsification without validating computational cost reduction.

To handle these problems, we introduce scheduled channel-wise sparsity (ssProp) during back-propagation, designed for seamless integration with PyTorch, i.e., simply replacing the built-in CNN module with our custom efficient CNN module, and compatibility with any machine or device. This flexibility allows our method to scale across various datasets and model architectures, as demonstrated in the experiment section where we apply it to a range of datasets from MNIST to the large-scale dataset of Imagenet-1k \cite{deng2009imagenet}, and from classification tasks to generation tasks using the most advanced generative model of denoising diffusion probabilistic models (DDPM) \cite{ho2020denoising}. Extensive experiments reveal that our method reduces computational resources required for the backward process by nearly 40\%, the most computation-intensive part, while also enhancing model performance by alleviating over-fitting similar to Dropout. Additionally, experiments show that our method is trustworthy to reduce energy consumption during the research and development (R\&D) of new AI models.

\section{Method}

\begin{figure*}
    \centering
    \includegraphics[width=0.8\textwidth]{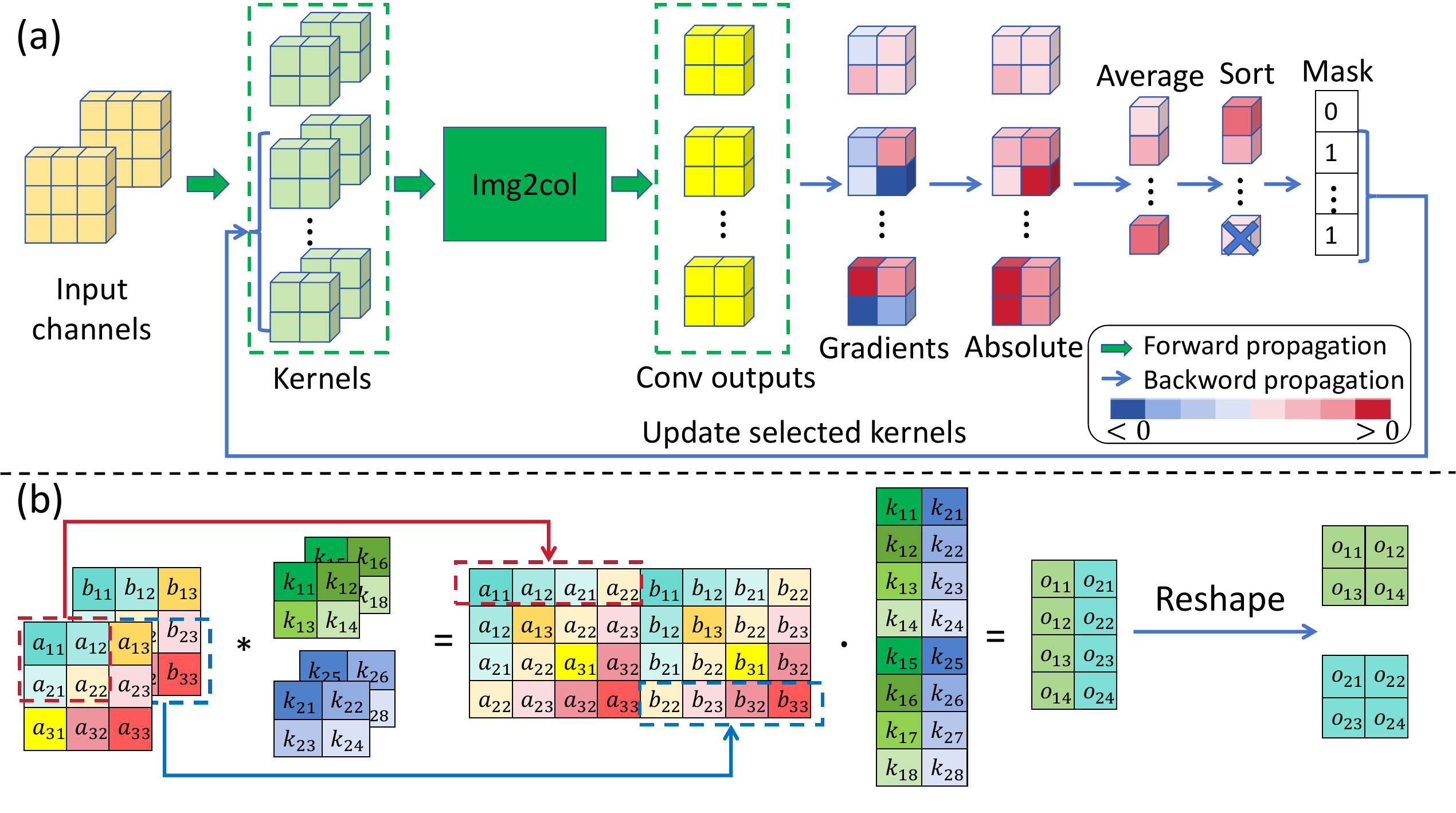}
    \caption{Overview of our method in CNNs. (a) Workflow in training one convolution layer. (b) Flowchart of the convolution using the img2col and col2img, demonstrated with input shape of (1, 2, 3, 3) and a kernel size of (2, 2, 2, 2).}
    \label{fig:img2col}
\end{figure*}

\subsection{Preliminaries}

We first introduce the forward and backward processes of CNN and then an efficient and effective computation acceleration method that inspires us to sparsify CNN back-propagation gradient maps.

\subsubsection{Forward Propagation}
We consider a single convolution layer with input $X$ in the shape of $(Bt, C_{in}, H_{in}, W_{in})$ and output $Y$ in the shape of $(Bt, C_{out}, H_{out}, W_{out})$, where $Bt$ is the batch size, $C_{in}$ and $C_{out}$ are the number of input and output channels, respectively; $H_{in}$, $W_{in}$, $H_{out}$, and $W_{out}$ are the height and width of the input feature maps, and the height and width of the output feature maps, respectively. Let $W$ be the convolution weights (filters) of shape $(C_{out}, C_{in}, K, K)$ and $B$ be the bias vector of shape $(C_{out})$, where $K$ represents the kernel size. Without losing generality, we assume the following for simplicity :

\begin{enumerate}
\item The kernel height equals the kernel width, denoted as $K$.
\item The stride of the convolution operation is denoted by $s$, which is a variable in this context.
\item The padding parameter $p$, dilation parameter $d$, and groups parameter $g$ are set to 0, 1, and 1 (default values), respectively.
\end{enumerate}

These assumptions are made for better explanation in this paper, but formulas can be easily extended to handle different parameters. In our implementation, parameters are not fixed and can be any values as mainstream deep learning frameworks support.

Following these assumptions, the forward pass is given by:

\begin{equation}
Y_{b, q, i, j} = \sum_{p=0}^{C_{in}-1} \sum_{m=0}^{K-1} \sum_{n=0}^{K-1} W_{q, p, m, n} \cdot X_{b, p, s \cdot i+m, s \cdot j+n} + B_q
\end{equation}

\noindent where $Y$ is the output feature map of shape $(Bt, C_{out}, H_{out}, W_{out})$, and $B_q$ is the bias for the $q^{th}$ output channel.

\subsubsection{Backward Propagation}
The back-propagation formulas for this convolution operation are:

\begin{enumerate}
\item Gradient w.r.t. single output element $Y_{b, q, i, j}$ from the output feature maps $Y$:
\begin{equation}
\frac{\partial L}{\partial Y_{b, q, i, j}}
\label{eq:grad_y}
\end{equation}

Where $L$ indicates the loss.

\item Gradient w.r.t. single weight element $W_{q, p, m, n}$ from the weights $W$ ($\frac{\partial L}{\partial W_{q, p, m, n}}$):
\begin{equation}
\sum_{b=0}^{Bt-1} \sum_{i=0}^{H_{out}-1} \sum_{j=0}^{W_{out}-1} X_{b, p, i+s \cdot m, j+s \cdot n} \cdot \frac{\partial L}{\partial Y_{b, q, i, j}}
\label{eq:grad_w}
\end{equation}

\item Gradient w.r.t. single input element $X_{b, p, i, j}$ from the input feature maps $X$ ($\frac{\partial L}{\partial X_{b, p, i, j}}$):
\begin{equation}
\sum_{q=0}^{C_{out}-1} \sum_{m=0}^{K-1} \sum_{n=0}^{K-1} \frac{\partial L}{\partial Y_{b, q, i-m, j-n}} \cdot W_{q, p, m, n}
\label{eq:grad_x}
\end{equation}

\item Gradient w.r.t. single bias element $B_q$ from biases $B$ ($\frac{\partial L}{\partial B_q}$):
\begin{equation}
\sum_{b=0}^{Bt-1} \sum_{i=0}^{H_{out}-1} \sum_{j=0}^{W_{out}-1} \frac{\partial L}{\partial Y_{b, q, i, j}} \cdot 1
\label{eq:grad_b}
\end{equation}

\end{enumerate}

\subsubsection{Img2col and Col2img}
Implementing forward and backward propagation in a pure 7-layer-loop manner is impractical due to its inefficiency and slow performance. Nowadays, many deep learning frameworks adopt various techniques to accelerate such computation-intensive processes, including the img2col and col2img algorithms. These algorithms have been explicitly implemented in frameworks like Caffe \cite{jia2014caffe} and are commonly used by other deep learning frameworks under the hood, especially in computation environments where, for example, GPU is unavailable.

An intuitive understanding of img2col is that it transforms the element-by-element multiplication of the convolution operation into matrix multiplication, enabling parallel computation and accelerating model training by introducing redundancy. As shown in Fig.~\ref{fig:img2col} (b), img2col stretches each convolution unit in the input $X$ (a 2D demonstration) into a single row, then concatenates all such rows together in the first dimension. Different data in a batch can also be stacked in a sequence for batch processing, finally resulting in a transformed $col\_X$ in the shape of $(Bt \cdot H_{out} \cdot W_{out}, C_{in} \cdot K \cdot K)$. Likewise, the weights (filters) $W$ can also be stretched in a similar way as Fig.~\ref{fig:img2col} (b) demonstrates, resulting in a transformed $col\_W$ in the shape of $(C_{in} \cdot K \cdot K, C_{out})$. By multiplying $col\_X$ with $col\_W$, we'll obtain the final convoluted results $col\_Y$, which can be reshaped to $(Bt, C_{out}, H_{out}, W_{out})$ to match the size of $Y$.

As for backward, the calculation of gradients of weights $W$ and input $X$ in Eq.~\ref{eq:grad_w} and Eq.~\ref{eq:grad_x} can be easily simplified as matrix multiplications between $col\_X$ and $col[\frac{\partial L}{\partial Y}]$, and between $col\_W$ and $col[\frac{\partial L}{\partial Y}]$, respectively, where the $col[\frac{\partial L}{\partial Y}]$ indicates the columnized form of $\frac{\partial L}{\partial Y}$. The col2img algorithm is the reverse process of img2col, transforming the calculated gradients of $col\_X$ in the shape of $(Bt \cdot H_{out} \cdot W_{out}, C_{in} \cdot K \cdot K)$ back to the shape of input $X$, i.e. $(Bt, C_{out}, H_{out}, W_{out})$, by summing the gradients that correspond to the same position in $X$.

\subsubsection{Backward Computation}

To quantify energy efficiency, we approximately measure the FLOPs of convolution back-propagation, which consumes the majority of computational resources. Although almost negligible, we account for the FLOPs of BatchNorm and Dropout modules. In the columnized form of convolution, the backward FLOPs for a single convolution operation can be easily calculated as:

\begin{equation}
(Bt \cdot H_{out} \cdot W_{out}) (4 C_{in} \cdot K^2 + 1) C_{out}
\label{eq:flops_conv}
\end{equation}

Independent of convolution operations, the backward FLOPs of BatchNorm and Dropout modules can be calculated as demonstrated in Eq.~\ref{eq:flops_batchnorm} and Eq.~\ref{eq:flops_dropout}, respectively.

\begin{equation}
12 (Bt \cdot H_{in} \cdot W_{in} \cdot C) + 10 C
\label{eq:flops_batchnorm}
\end{equation}

\begin{equation}
2 (Bt \cdot H_{in} \cdot W_{in} \cdot C)
\label{eq:flops_dropout}
\end{equation}

\subsection{Scheduled Sparse Back Propagation}

To introduce sparsity to gradient maps during back-propagation, previous works \cite{wei2017minimal,sun2018training,raihan2020sparse} typically set a majority of matrix elements to zero, which enables skipping most matrix multiplication operations to reduce computational costs. However, such methods depend on specific sparsity patterns such as 2-out-of-4 non-zero pattern \cite{NVIDIA2020} and specialized instructions support from hardware like NVIDIA A100. While training on GPUs that do not support sparsity acceleration, which is much more common nowadays, these methods usually do not lead to desirable sparsity training acceleration and computational savings.
To promote general sparse training on any machines and devices, we seek more structured matrix sparsity that doesn't rely on hardware support to reduce computational costs. Specifically, we introduce channel-wise sparsity similar to works using parametric channel-level sparsification \cite{zhou2021efficient}. Instead of introducing the regularization term into loss functions during training, we utilize gradients w.r.t top-K important output channels during backward and propose gradients selection schedulers to boost sparse training performance.
As illustrated in Fig.~\ref{fig:img2col} (a), with output gradients, we first compute the absolute values and average them spatially. This results in a vector where each element represents the magnitude, or 'importance,' of the corresponding output channel. Channels with larger importance are prioritized as they contribute more significantly to the gradients of inputs and weights/biases. We then sort this importance vector to create a mask that identifies the indices of the top K important channels. Gradients of these top-K channels are retained for calculating downstream gradients, while others are dropped. After gradients selection, we end up with a shrunk $col[\frac{\partial L}{\partial Y}]'$ in the shape of $(Bt \cdot H_{out} \cdot W_{out}, C_{out}')$, where the $C_{out}'$ denotes the selected top-K output channels. The gradients of columnized inputs $col\_X$ and weights $col\_W$ can be calculated by multiplication between $col\_W$ and $col[\frac{\partial L}{\partial Y}]'$, and between $col\_X$ and $col[\frac{\partial L}{\partial Y}]'$, respectively.

Previous sparsification-based approaches often compromise model performance as the drop rate increases. To address this issue, we draw inspiration from learning rate schedulers and propose the use of drop schedulers during sparse training to improve model performance. Specifically, we implement an 80\% sparsification strategy (discarding 80\% channels of output gradients) on a periodic basis across CNN layers. This involves training the model normally in epochs 1, 3, 5, and so on while applying the 80\% drop rate sparsification during epochs 2, 4, 6, etc., and similarly in subsequent epochs.

Below, we conduct a sensitivity analysis to evaluate different sparsification strategies and validate the specifics of our approach. We focus on the following key aspects: the sparsified dimensions (channel, width, height, or a combination), the gradients selection method (Top-K or random), the drop rate, and the drop schedulers.

\subsubsection{Sensitivity Analysis}

\begin{figure}
    \centering
    \includegraphics[width=1.0\linewidth]{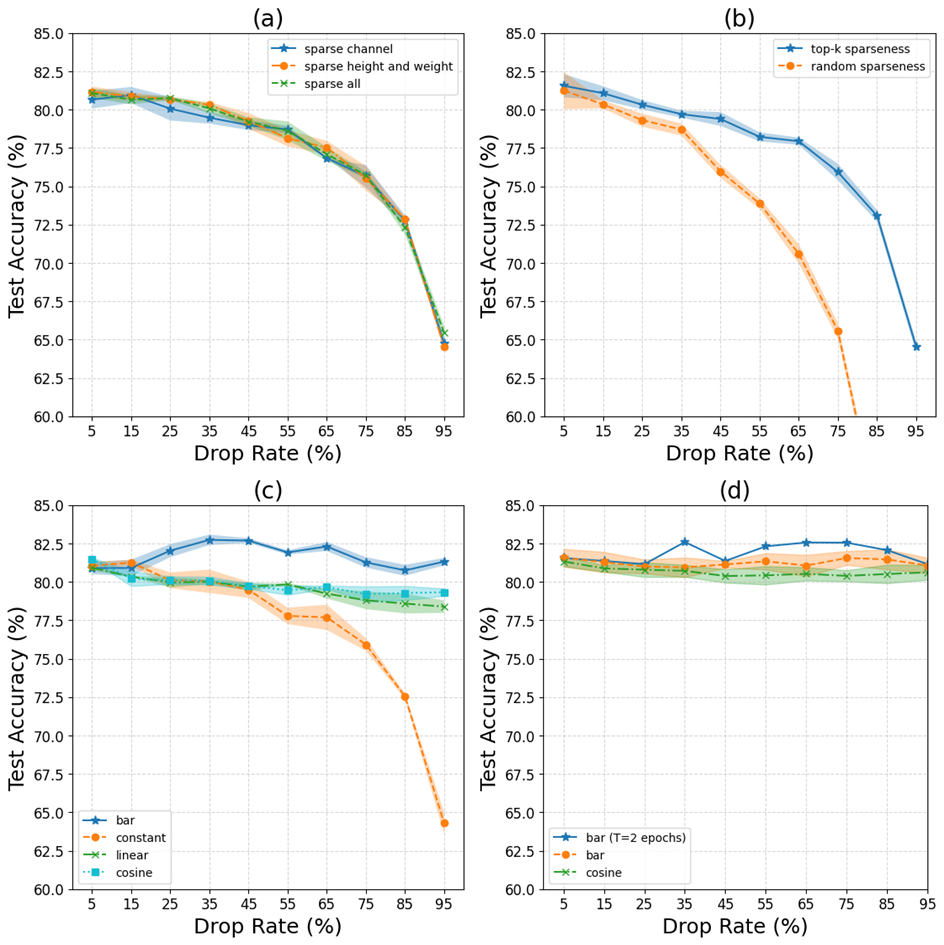}
    \caption{Sensitivity analysis. (a) Sparisified dimensions. (b) Gradients selection. (c) \& (d) Sparsification schedulers.}
    \label{fig:sensitivity_analysis}
\end{figure}

We run experiments on CIFAR10 using ResNet18 with different drop rates ranging from 5 percent to 95 percent. All experiments are conducted 3 times and reported with the mean and standard deviation values unless explicitly specified otherwise. The sparsification is applied to all convolution layers within the model.

Fig.~\ref{fig:sensitivity_analysis} plots the impact of various dropping behaviors on CIFAR10 test accuracy. We first explore three types of ways to sparsify the gradient maps: sparse-channel, sparse-height-and-weight, and sparse-all. These methods have exactly the same sparsification process as demonstrated in Fig.~\ref{fig:img2col} (a), with a difference in averaging dimensions. The sparse-channel method averages gradients on dimensions of $Bt$, $H_{out}$, and $W_{out}$. The sparse-height-and-weight and sparse-all methods have similar processes with averaging dimensions of $Bt$ and $C_{out}$, and $Bt$, respectively.
Fig.~\ref{fig:sensitivity_analysis} (a) illustrates decreasing test accuracy as the drop rate increases, with minor differences among the three dropping behaviors across different drop rates. In practice, the sparse-channel mode is more practical due to its concise implementation and efficient integration with PyTorch, motivating us to continue with this mode in further experiments. To determine the necessity of sorting and retaining the largest, we compare the top-k method with random dropping in Fig.~\ref{fig:sensitivity_analysis} (b), which shows that the model performance degrades much faster with random sparsity.

In Fig.~\ref{fig:sensitivity_analysis} (c), we explore how drop schedulers affect the model convergence. We apply drop schedulers that gradually increase the drop rate from 0\% to a specified rate (the X-axis) in linear, cosine, and bar patterns from the first epoch to the last epoch. For instance, with a target drop rate of 55\%, we conduct four experiments independently using bar, linear, cosine schedulers, and a constant mode. In each experiment, the drop rate starts at 0\% in the first epoch and gradually increases to 55\% by the final epoch, according to the specific schedulers. The bar scheduler functions like a step function, maintaining a 0\% drop rate for the first half of the training before abruptly increasing to the target rate in the second half. The constant mode serves as a baseline, keeping the drop rate fixed throughout the entire training process. The results indicate that schedulers can significantly improve model performance compared to the constant mode, even with a high drop rate of 95\%. Among the three types of schedulers, the bar scheduler performs the best with drop rates ranging from 25\% to 95\% (Fig.~\ref{fig:sensitivity_analysis} (c) blue line).

Next, we compare various scheduler periods, ranging from 30 iterations to 300 iterations in increments of 30 (with 300 iterations nearly representing an epoch) and a 2-epoch-long period, which alternates between 0\% drop rate for one epoch and a certain drop rate for the next epoch. In this analysis, we conduct each experiment only once. The mean and the standard deviation for bar and cosine schedulers in Fig.~\ref{fig:sensitivity_analysis} (d) are calculated among different periods of iterations. The bar scheduler with a period of 2 epochs is plotted separately. The results indicate that the period length does not significantly impact performance when it ranges from 30 to 300 iterations, as evidenced by the narrow shadow area (low standard deviation) in Fig.~\ref{fig:sensitivity_analysis} (d). Moreover, the bar scheduler with a period of 2 epochs outperforms the others by more than one standard deviation. 

To optimize computational efficiency while preserving model performance, we implement sparsification along the channel dimension by discarding 80\% of the smallest gradients, utilizing a bar scheduler with a period of 2 epochs.

\subsubsection{Drop Rate Lower Bound}

Because of the necessity of finding the smallest/largest gradients, we determine the minimum drop rate required to save computation without introducing additional overhead. In computing FLOPs for CNN, BatchNorm, and Dropout modules as shown in Eq.~\ref{eq:flops_conv}, Eq.~\ref{eq:flops_batchnorm}, and Eq.~\ref{eq:flops_dropout}, each operation of Addition, Subtraction, Multiplication, or Division is counted as one FLOP. The process of finding the smallest/largest gradients includes a summation operation across the batch, height, and width dimensions and a sorting operation along the channel dimension. The sorting operation only involves comparisons and does not generate floating-point computation, but the summation operation introduces additional $(Bt \cdot H_{out} \cdot W_{out} - 1) \cdot C_{out}$ FLOPs.

Approximately, to save back-propagation computation, the drop rate $D$ needs to satisfy the following inequality:

\begin{flalign}
&\underbrace{(4MN + M) C_{out}}_{\text{CNN FLOPs}} > \underbrace{[(4MN + M) (1 - D) + M] C_{out}}_{\text{CNN FLOPs w/ ssProp}}&
\label{eq:flops_inequality}
\end{flalign}

\noindent where $M = (Bt \cdot H_{out} \cdot W_{out})$ and $N = (C_{in} \cdot K^2)$. By solving this inequality, the lower bound of the drop rate is defined as:

\begin{equation}
D > \frac{1}{4 N + 1} = \frac{1}{4 C_{in} \cdot K^2 + 1}
\label{eq:flops_lower_bound}
\end{equation}

Assuming $K \geq 3$ and $C_{in} \geq 1$, we have:

\begin{equation}
\frac{1}{4 C_{in} \cdot K^2 + 1} \leq 0.027027... \approx 3\%
\label{eq:flops_lower_bound}
\end{equation}

Therefore, as long as the drop rate is set to at least 3\%, the training computation will theoretically be reduced. Given our strategy of using the bar scheduler with a drop rate of 80\%, the average drop rate throughout the training process will be approximately 40\%, significantly exceeding the lower bound required to achieve computational savings.

\section{Experiments}

\begin{table}[t]
\centering
\resizebox{1.\columnwidth}{!}{
\begin{tabular}{c c c c c}
\toprule
    Dataset & Total Number & Train/Val/Test & Image Size & Class Number \\
\midrule
MNIST & 70,000 & 48,000/12,000/10,000 & (1, 28, 28) & 10 \\
\midrule
FashionMNIST & 70,000 & 48,000/12,000/10,000 & (1, 28, 28) & 10 \\
\midrule
CIFAR10 & 60,000 & 40,000/10,000/10,000 & (3, 32, 32) & 10 \\
\midrule
CIFAR100 & 60,000 & 40,000/10,000/10,000 & (3, 32, 32) & 100 \\
\midrule
CelebA & 202,599 & 162,770/19,867/19,962 & (3, 64, 64) & 40 \\
\midrule
Imagenet-1k & 1,431,167 & 1,281,167/50,000/100,000 & (3, 224, 224) & 1000 \\
\bottomrule
\end{tabular}
}
\caption{Datasets for classification and generation tasks.}
\label{tab:dataset}
\end{table}

\subsection{Implementation}

The img2col algorithm allows us to discard unwanted gradients and compute FLOPs savings conveniently, but it is quite impractical for parallel training on GPUs, limiting our method to small models and datasets. Fortunately, we can integrate our method into PyTorch by leveraging its built-in back-propagation implementation for fast training on any machine. 
Our implementation includes both the img2col version and PyTorch built-in backward version, which can be shifted as needed under various computing environments. 

To validate our method, we run ResNet-18 and ResNet-50 \cite{he2016deep} on the MNIST, FashionMNIST, CIFAR10, CIFAR100, CelebA, and Imagenet-1k datasets \cite{lecun1998gradient,xiao2017fashion,krizhevsky2009learning,liu2018large,deng2009imagenet}, as detailed in Table~\ref{tab:dataset}. The CelebA dataset, which contains 40 binary labels per image, initially has image dimensions of (3, 218, 178). These images are resized to (3, 64, 64) in our experiments. For the Imagenet-1k dataset, we address varying image sizes by first resizing the images to (3, 256, 256) and then cropping the center to (3, 224, 224). To train the ResNet models on these datasets, we normalize each image using the mean and standard deviation specific to the respective dataset. To thoroughly evaluate our method, we assess its performance on both classification and generation tasks. 

\begin{table}[t]
\centering
\resizebox{1.\columnwidth}{!}{
\begin{tabular}{c c c c c c}
\toprule
    Task & Dataset & Model & \makecell{Learning \\ Rate} & Epoch & Batch Size \\
\midrule
\multirow{6}{*}{Classification} & MNIST & ResNet-18/50 & 2e-4 & 50/50 & 128/128 \\
\cmidrule(r){2-6}
 & FashionMNIST & ResNet-18/50 & 2e-4 & 50/50 & 128/128 \\
\cmidrule(r){2-6}
 & CIFAR10 & ResNet-18/50 & 2e-4 & 50/250 & 128/128 \\
\cmidrule(r){2-6}
 & CIFAR100 & ResNet-18/50 & 2e-4 & 50/250 & 128/128 \\
\cmidrule(r){2-6}
 & CelebA & ResNet-18/50 & 2e-4 & 50/50 & 128/32 \\
\cmidrule(r){2-6}
 & Imagenet-1k & ResNet-18/50 & 2e-4 & 50/50 & 32/16 \\
\bottomrule
\end{tabular}
}
\caption{Training hyperparameters for classification tasks.}
\label{tab:classification_hyperparameter}
\end{table}

\begin{table}[t]
\centering
\resizebox{1.\columnwidth}{!}{
\begin{tabular}{c c c c c c c}
\toprule
    Task & Dataset & Model & \makecell{Learning \\ Rate} & Timesteps & Epoch & Batch Size \\
\midrule
\multirow{3}{*}{Generation} & MNIST & DDPM & 1e-3 & 200 & 300 & 128 \\
\cmidrule(r){2-7}
 & FashionMNIST & DDPM & 1e-3 & 200 & 500 & 128 \\
\cmidrule(r){2-7}
 & CelebA & DDPM & 2e-4 & 1000 & 200 & 128 \\
\bottomrule
\end{tabular}
}
\caption{Training hyperparameters for generation tasks.}
\label{tab:generation_hyperparameter}
\end{table}

The training hyperparameters for the classification and generation tasks are shown in Table~\ref{tab:classification_hyperparameter} and Table~\ref{tab:generation_hyperparameter}, respectively. All classification tasks are trained with the Adam optimizer with betas (0.9, 0.999), while all generation tasks are trained using the AdamW optimizer with default parameters. No learning rate schedulers and augmentation techniques are used in the training of any models, and all models are initialized with Kaiming Initialization \cite{he2015delving} before training. For the classification tasks in Table~\ref{tab:classification}, all experiments, except for ImageNet-1k, are conducted three times with different seeds and reported with average accuracy. The generation tasks in Table~\ref{tab:generation} are conducted once with the same seed. All experiments are conducted on Ubuntu 20.04.5 LTS (Focal Fossa) with the PyTorch 2.3.0 built-in backward version for efficiency, where the computing resources include a single 24G NVIDIA RTX A5000 GPU with AMD Ryzen Threadripper 3960X 24-Core Processor.

To compute the backward FLOPs consumption of ResNet and DDPM models during training, we calculate the FLOPs for the convolution layers, BatchNorm layers, and Dropout layers using Eq.~\ref{eq:flops_conv}, Eq.~\ref{eq:flops_batchnorm}, and Eq.~\ref{eq:flops_dropout}, respectively. These three types of modules account for almost all FLOPs consumption. For DDPM, we use GroupNorm instead of BatchNorm, but we exclude it from the calculation since convolution modules dominate the FLOPs consumption up to 99.7\%.

\subsection{Results}

\begin{figure}
    \centering
    \includegraphics[width=1.0\linewidth]{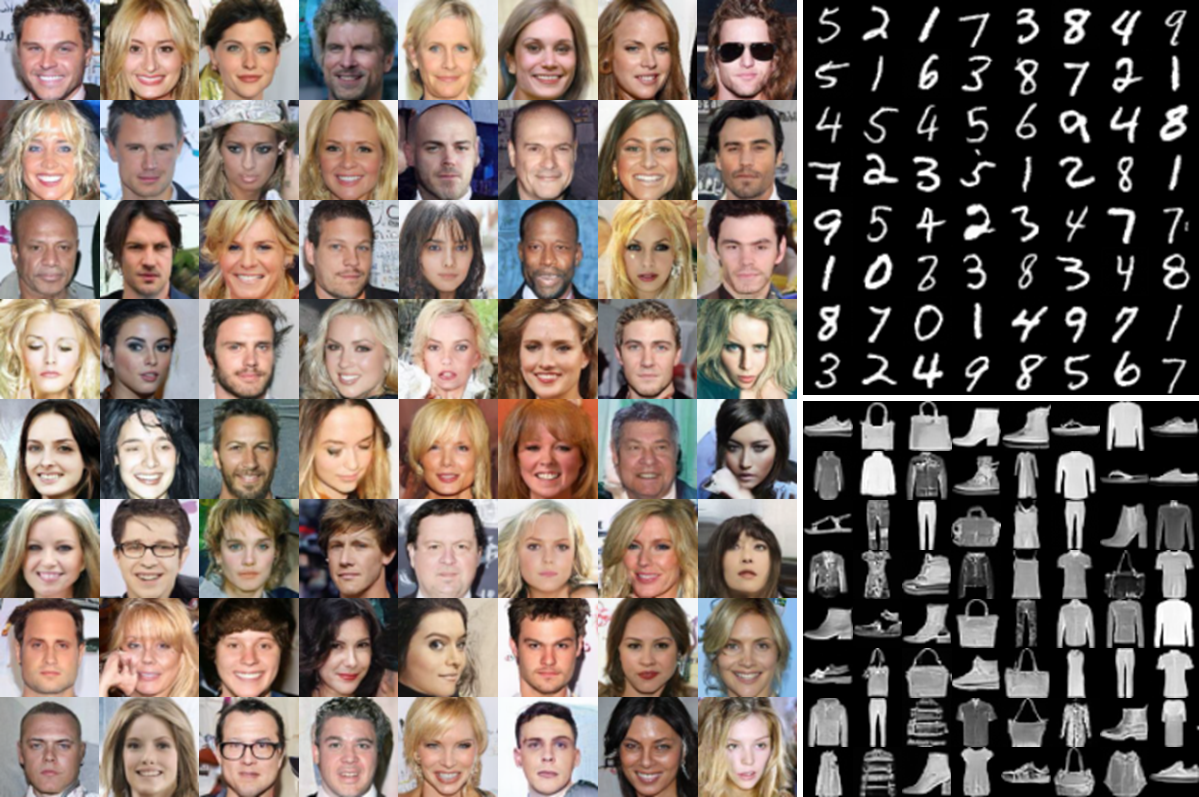}
    \caption{Sparsely trained DDPM-generated samples on MNIST, FashionMNIST, and CelebA.}
    \label{fig:ddpm_generation}
\end{figure}

Table~\ref{tab:classification} presents the classification performance of ResNet-18 and ResNet-50 on six different datasets, with data sizes ranging from tens of thousands to over one million and image scales from grayscale 28×28 to RGB 224×224. 
According to the FLOPs per iteration and total FLOPs consumption shown in the 3rd and 4th columns of Table~\ref{tab:classification}, our method reduces FLOPs during back-propagation by nearly 40\% compared to normal training. Despite these computational savings, our method achieves slightly better performance on almost all datasets compared to normally trained models. Additionally, using our method for back-propagation slightly decreases the training time. This may seem contradictive at first because the time consumption introduced by sorting is higher than the time saved from shrunk matrix multiplication, especially on GPUs with good computation capability. However, under the same GPU memory allocation, a large amount of shrunk matrices during back-propagation can significantly release memory occupation and speed up the data transfer between CPU and GPU, leading to faster training.

The superiority of our method is evident in generation tasks as well. Table~\ref{tab:generation} shows that our method not only reduces up to 40\% computation during back-propagation but also slightly improves model performance, as measured by FID, and reduces training time. Additional visual results of our method are illustrated in Fig.~\ref{fig:ddpm_generation}.

\begin{table}[t]
\centering
\resizebox{1.\columnwidth}{!}{
\begin{tabular}{c c c c c c}
\toprule
    Dataset & Model & \makecell{Est. FLOPs \\ (B./Iter.)} & \makecell{Total Est. FLOPs \\ (Quad.)} & \makecell{Train Time \\ (mins)} & \makecell{Test Acc \\ (\%)}\\
\midrule
\multirow{2}{*}{MNIST} & ResNet-18/50 & 234.10/540.06 & 5.49/12.66 & 11/29 & \textbf{99.56}/\textbf{99.47} \\
                       & ssProp-18/50 & \textbf{140.79}/\textbf{325.85} & \textbf{3.30}/\textbf{7.64} & \textbf{11}/\textbf{27} & 99.52/99.42 \\
\midrule
\multirow{2}{*}{FashionMNIST} & ResNet-18/50 & 234.10/540.06 & 5.49/12.66 & 12/30 & 92.41/92.64 \\
                              & ssProp-18/50 & \textbf{140.79}/\textbf{325.85} & \textbf{3.30}/\textbf{7.64} & \textbf{12}/\textbf{29} & \textbf{92.86}/\textbf{92.67} \\
\midrule
\multirow{2}{*}{CIFAR10} & ResNet-18/50 & 285.32/669.75 & 5.57/65.41 & 15/138 & 81.88/84.39 \\
                         & ssProp-18/50 & \textbf{171.61}/\textbf{404.18} & \textbf{3.35}/\textbf{39.47} & \textbf{14}/\textbf{133} & \textbf{81.89}/\textbf{85.44} \\
\midrule
\multirow{2}{*}{CIFAR100} & ResNet-18/50 & 285.32/669.75 & 5.57/65.41 & 36/135 & 54.15/54.20 \\
                          & ssProp-18/50 & \textbf{171.61}/\textbf{404.18} & \textbf{3.35}/\textbf{39.47} & \textbf{28}/\textbf{132} & \textbf{55.15}/\textbf{56.90} \\
\midrule
\multirow{2}{*}{CelebA} & ResNet-18/50 & 1,141.27/669.75 & 81.42/191.13 & 138/396 & 90.69/91.07 \\
                        & ssProp-18/50 & \textbf{686.43}/\textbf{404.18} & \textbf{48.97}/\textbf{115.34} & \textbf{120}/\textbf{390} & \textbf{91.23}/\textbf{91.20} \\
\midrule
\multirow{2}{*}{ImageNet-1k (top-1)} & ResNet-18/50 & 3,495.14/4,102.22 & 7,269.71/17,064.82 & 7,980/20,400 & 60.95/65.39 \\
                       & ssProp-18/50 & \textbf{2,102.19}/\textbf{2,475.60} & \textbf{4,372.45}/\textbf{10,298.23} & \textbf{7,200}/\textbf{19,200} & \textbf{61.16}/\textbf{65.96} \\
\midrule
\multirow{2}{*}{ImageNet-1k (top-5)} & ResNet-18/50 & 3,495.14/4,102.22 & 7,269.71/17,064.82 & 7,980/20,400 & 83.36/86.36 \\
                       & ssProp-18/50 & \textbf{2,102.19}/\textbf{2,475.60} & \textbf{4,372.45}/\textbf{10,298.23} & \textbf{7,200}/\textbf{19,200} & \textbf{83.39}/\textbf{86.61} \\
\bottomrule
\end{tabular}
}
\caption{Classification tasks performance comparison between models trained normally (ResNet-) and with our method (ssProp-). We report the estimated (Est.) backward FLOPs per iteration in Billions (B./Iter.) and the total backward FLOPs during training in Quadrillions (Quad.)}
\label{tab:classification}
\end{table}

\begin{table}[t]
\centering
\resizebox{1.\columnwidth}{!}{
\begin{tabular}{c c c c c c}
\toprule
    Dataset & Model & \makecell{Est. FLOPs \\ (B/Iter)} & \makecell{Total Est. FLOPs \\ (Quad.)} & \makecell{Train Time \\ (mins)} & FID \\
\midrule
\multirow{2}{*}{MNIST} & DDPM & 65.05 & 9.15 & \textbf{100} & 0.06 \\
                       & ssProp-DDPM & \textbf{39.06} & \textbf{5.49} & 106 & \textbf{0.04} \\
\midrule
\multirow{2}{*}{FashionMNIST} & DDPM & 65.05 & 15.24 & \textbf{168} & 0.32 \\
                              & ssProp-DDPM & \textbf{39.06} & \textbf{9.15} & 176 & \textbf{0.18} \\
\midrule
\multirow{2}{*}{CelebA} & DDPM & 11,696.81 & 3,337.92 & 3,780 & 5.98 \\
                         & ssProp-DDPM & \textbf{7,018.96} & \textbf{2,003.00} & \textbf{3,208} & \textbf{5.97} \\
\bottomrule
\end{tabular}
}
\caption{Generation tasks performance comparison.}
\label{tab:generation}
\end{table}

\subsection{Analysis of Key Properties}
Although our method shows promising results in both classification and generation tasks, it is also important to investigate the following key properties: 
\begin{enumerate}
\item Overfitting Prevention (Q1): Can it prevent overfitting like Dropout?
\item Comparison to Models with Similar FLOPs (Q2): Can it outperform a model with similar FLOPs consumption?
\item Model Reliability (Q3): is sparsely trained model trustworthy?
\end{enumerate}

All experiment settings in this section are the same as stated previously, except as described below. For Q1, models with Dropout \cite{srivastava2014dropout} are trained for 4000 epochs on CIFAR10 and 1800 epochs on CIFAR100 due to the slower convergence caused by the Dropout layer. For the experiments on Q1 and Q2, as shown in Table~\ref{tab:dropout} and Table~\ref{tab:smaller_model}, respectively, all experiments are conducted two times with different seeds and reported with average accuracy. As for Q3, the CNNs are trained once for 50 epochs on CIFAR100 using the Adam optimizer.

\subsubsection{Overfitting Prevention (Q1)}

Dropout is a regularization technique to prevent overfitting, which sets partial inputs to zeros during the forward process. During backward, those zeroed elements contribute zero gradients, helping prevent the network from becoming overly reliant on specific elements and enhancing model generalization. Our method aims to reduce computational expense during training without compromising performance, distinguishing itself from Dropout, which even increases computational demands in both forward and backward passes. As shown in Table~\ref{tab:dropout}, adding Dropout layers leads to higher training costs in terms of time and FLOPs due to slower convergence, resulting in higher test accuracy than the baseline. In contrast, our method also improves test accuracy over ResNet-50 while saving 40\% FLOPs during backward, leading to comparable performance to Dropout with only around 4\% and 8\% backward FLOPs consumption on CIFAR10 and CIFAR100, respectively. This indicates that our method saves computation with the benefit of preventing overfitting.

Additionally, experiments show that our method prevents overfitting in a different way compared to the Dropout. The last two rows for each dataset in Table~\ref{tab:dropout} illustrate that combining Dropout with our method results in even higher test accuracy than using Dropout alone. For CIFAR10 and CIFAR100, the 0.4+0.4 and 0.2+0.2 modes perform best, respectively. This is expected since the ResNet-50 model is less likely to overfit on the more complex dataset of CIFAR100, requiring less sparsification compared to CIFAR10. Even the 0.2+0.2 mode on CIFAR10 and 0.4+0.4 mode on CIFAR100 achieve comparable performance to using only Dropout while also saving 40\% of computation. These improvements highlight the effectiveness of our approach to integrate with Dropout to enhance model performance and reduce computational costs.

\begin{table}[t]
\centering
\resizebox{1.\columnwidth}{!}{
\begin{tabular}{c c c c c c}
\toprule
    Dataset & Method (Drop Rate) & \makecell{Est. FLOPs \\ (B/Iter)} & \makecell{Total Est. FLOPs \\ (Quad.)} & \makecell{Train Time \\ (mins)} & \makecell{Test Acc \\ (\%)} \\
\midrule
\multirow{4}{*}{CIFAR10} & ResNet-50 (0) & 669.75 & \underline{65.41} & \underline{140} & 84.44 \\
\cmidrule(r){2-6}
                         & w/ Dropout (0.4) & 671.51 & 1,046.49 & 2,378 & \underline{87.80} \\
\cmidrule(r){2-6}
                         & w/ ssProp (0.4) & \textbf{404.18} & \textbf{39.47} & \textbf{130} & 85.21 \\
\cmidrule(r){2-6}
                         & w/ Both (0.2 + 0.2) & 538.83 & 839.18 & 2,533 & 87.24 \\
\cmidrule(r){2-6}
                         & w/ Both (0.4 + 0.4) & \underline{405.94} & 631.53 & 2,175 & \textbf{87.89} \\
\midrule
\multirow{4}{*}{CIFAR100} & ResNet-50 (0) & 669.75 & \underline{65.41} & \underline{158} & 54.86 \\
\cmidrule(r){2-6}
                          & w/ Dropout (0.4) & 671.51 & 470.92 & 1,295 & \underline{58.11} \\
\cmidrule(r){2-6}
                          & w/ ssProp (0.4) & \textbf{404.18} & \textbf{39.47} & \textbf{153} & 56.17 \\
\cmidrule(r){2-6}
                          & w/ Both (0.2 + 0.2) & 538.83 & 377.63 & 1,155 & \textbf{58.84} \\
\cmidrule(r){2-6}
                          & w/ Both (0.4 + 0.4) & \underline{405.94} & 284.19 & 1,076 & 57.24 \\
\bottomrule
\end{tabular}
}
\caption{Performance comparison of ResNet-50 between dropout and our method.}
\label{tab:dropout}
\end{table}

\subsubsection{Comparison to Models with Similar FLOPs (Q2)}

\begin{table}[t]
\centering
\resizebox{1.\columnwidth}{!}{
\begin{tabular}{c c c c c c}
\toprule
    Dataset & Method & \makecell{Est. FLOPs \\ (B/Iter)} & \makecell{Total Est. FLOPs \\ (Quad.)} & \makecell{Train Time \\ (mins)} & \makecell{Test Acc \\ (\%)} \\
\midrule
\multirow{4}{*}{CIFAR10} & ResNet-50 & 669.75 & 65.41 & 140 & 84.44 \\
\cmidrule(r){2-6}
                         & ssProp-50 & \underline{404.18} & \underline{39.47} & 130 & \underline{85.21} \\
\cmidrule(r){2-6}
                         & ResNet-26 & 440.19 & 42.99 & \underline{81} & 85.10 \\
\cmidrule(r){2-6}
                         & ssProp-26 & \textbf{264.64} & \textbf{25.84} & \textbf{71} & \textbf{85.24} \\
\midrule
\multirow{4}{*}{CIFAR100} & ResNet-50 & 669.75 & 65.41 & 158 & 54.86 \\
\cmidrule(r){2-6}
                          & ssProp-50 & \underline{404.18} & \underline{39.47} & 153 & 56.17 \\
\cmidrule(r){2-6}
                          & ResNet-26 & 440.19 & 42.99 & \underline{71} & \underline{56.27} \\
\cmidrule(r){2-6}
                          & ssProp-26 & \textbf{264.64} & \textbf{25.84} & \textbf{70} & \textbf{57.55} \\
\bottomrule
\end{tabular}
}
\caption{Performance comparison between sparsely trained ResNet-50 (ours) and normally trained smaller models.}
\label{tab:smaller_model}
\end{table}

To compare our method to a normally trained model with similar FLOPs consumption, we design a new ResNet model called ResNet-26, featuring \textit{BasicBlock} layers in the configuration (2, 3, 5, 2). This model is compared to a sparsely trained ResNet-50, which uses \textit{Bottleneck} layers in the configuration (3, 4, 6, 3). Both models exhibit similar backward FLOPs consumption during training. As shown in Table~\ref{tab:smaller_model}, we run both ResNet-26 and ResNet-50 in two different modes. The results demonstrate that the sparsely trained ResNet-50 does have equivalent test accuracy to the normally trained ResNet-26. This raises a key question: why not simply downscale the model by 40\% instead of applying sparsification? The answer lies in the tendency of the larger model (ResNet-50) to overfit the dataset more than the smaller model (ResNet-26). A larger model is still preferable when dealing with diverse and complex datasets like Imagenet-1k. Either way, by introducing sparsification, both ssProp-50 and ssProp-26 (i.e., sparsely trained ResNet-50 and ResNet-26) outperform their counterparts (i.e., normally trained ResNet-50 and ResNet-26) by a considerable margin, demonstrating the superiority of our method.

\subsubsection{Model Reliability  (Q3)} A reasonable question arises when developing new AI models: can we really trust the sparsely trained model? In other words, will the model that performs best under normal training also perform best while sparsely trained? If so, this approach could make AI research and development more environmentally sustainable by substantially lowering energy consumption and reducing the carbon footprint.

\begin{figure}
    \centering
    \includegraphics[width=1.0\linewidth]{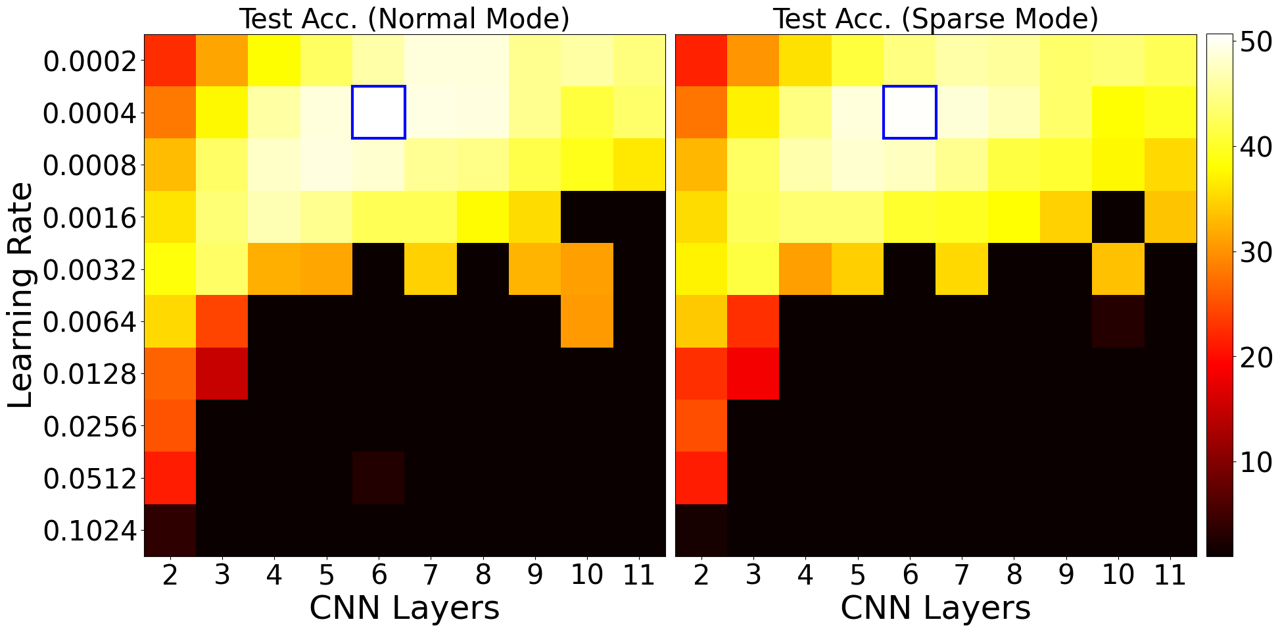}
    \caption{Test accuracy patterns of sparsely and normally trained CNN models on CIFAR100.}
    \label{fig:heatmap}
\end{figure}

To conduct a large-scale hyperparameters searching experiment, we adopt a simple CNN architecture with a few convolutional layers followed by a fully connected layer for classification. The number of CNN layers in the model ranges from 2 to 11, and the learning rate ranges from 0.0002 to 0.1024, increasing by a factor of 2. We run models on CIFAR100 and compare sparsely and normally trained models by observing the test accuracy patterns. As Fig.~\ref{fig:heatmap} presents, the overall performance patterns are very similar between the normal and sparse modes, except for some cases with extremely unsuitable learning rates. This is particularly evident in the blue-highlighted area in Fig.~\ref{fig:heatmap}, where the best-performing models in both modes share the same hyperparameters of 6 CNN layers and a learning rate of 4e-4. This indicates that our method is reliable to use during the development and research phase to find the best hyperparameters while saving computational resources.

\section{Conclusion}

In this work, we propose a novel sparsification method that saves nearly 40\% of computational resources during back-propagation while slightly improving model performance by preventing overfitting. Our method flexibly scales up to large-scale datasets like Imagenet-1k and performs well on both discriminative and generative tasks. It is efficient and compatible with many deep learning techniques, such as BatchNorm, GroupNorm, and Dropout, without requiring hardware sparsification acceleration support from GPUs like A100. To facilitate efficient AI development and reduce the carbon footprint, our method can be utilized during the R\&D phases of AI technologies without compromising model performance and hyperparameters searching.

Although this paper has made progress and improvements, there is still much to be investigated in the future: 
(1) Other techniques to enhance model performance, such as warm-up before sparsification, different drop rates and schedulers across layers within the model, etc; (2) Other properties of our sparsification, like sensitivity to data noise or feature distortion; (3) Improvement of sparsification by getting rid of sorting.
Additionally, it is crucial to extend our method to MLPs and Transformers, which are the foundations of large language models, to facilitate efficient training in the natural language processing field.

\section{Acknowledgments}

This work was supported by the National Institute of Health (NIH) under grants R01EB022744, RF1AG077578,  RF1AG064584, RF1AG084072, and U19AG078109.

\bibliography{aaai24}

\end{document}